# CIFAR10 to Compare Visual Recognition Performance between Deep Neural Networks and Humans

Tien Ho-Phuoc
The University of Danang – University of Science and Technology
hptien@dut.udn.vn

**Abstract**

*Visual object recognition plays an essential role in human daily life. This ability is so efficient that we can recognize a face or an object seemingly without effort, though they may vary in position, scale, pose, and illumination. In the field of computer vision, a large number of studies have been carried out to build a human-like object recognition system. Recently, deep neural networks have shown impressive progress in object classification performance, and have been reported to surpass humans. Yet there is still lack of thorough and fair comparison between humans and artificial recognition systems. While some studies consider artificially degraded images, human recognition performance on dataset widely used for deep neural networks has not been fully evaluated. The present paper carries out an extensive experiment to evaluate human classification accuracy on CIFAR10, a well-known dataset of natural images. This then allows for a fair comparison with the state-of-the-art deep neural networks. Our CIFAR10-based evaluations show very efficient object recognition of recent CNNs but, at the same time, prove that they are still far from human-level capability of generalization. Moreover, a detailed investigation using multiple levels of difficulty reveals that easy images for humans may not be easy for deep neural networks. Such images form a subset of CIFAR10 that can be employed to evaluate and improve future neural networks.*

## 1. Introduction

Visual object recognition is of great importance for humans to interact with each other and the natural world. We possess a huge ability of visual recognition as we can almost effortlessly recognize objects encountered in our life such as animals, faces, and food. Especially, humans can easily recognize an object even though it may vary in position, scale, pose, and illumination. Such ability is called core object recognition, and is carried out through the ventral stream in the human visual system [1]. Visual object recognition has long been considered as a privilege of biological systems.

In the field of computer vision, many studies have tried to build systems able to imitate humans' object recognition ability. In spite of several decades of effort, machine visual recognition was far from human performance. Yet, since past few years, machine performance has been dramatically improved thanks to the reemergence of convolutional neural networks (CNN) and deep learning [2, 3, 4, 5, 6], and thus even surpasses human performance [7, 8]. Like traditional neural networks, which are inspired by biological neural systems, the architecture of CNNs for object recognition is feedforward and consists of several layers in a hierarchical manner. Particularly, some works reveal hierarchical correspondence between CNN layers and those in the human object recognition system [9].

Object recognition performance of deep neural networks is often measured on datasets commonly used in the field such as ImageNet, CIFAR100, and CIFAR10. However, according to our best knowledge, these datasets have not been thoroughly tested with human subjects in order to make a detailed comparison between artificial vision systems and humans. In fact, while some CNNs are supposed to outperform humans in object recognition, such comparison is usually based on the human performance provided in [10], which is evaluated on the ImageNet dataset. Yet in [10] human subjects did not recognize the test images in the same way as CNNs; there were only two persons participating in the recognition experiment: one person classified 1500 images and the other 258 images out of 100000 test images. Meanwhile, CNNs are trained to predict all of this test dataset.

In order to obtain a thorough and fair comparison in object recognition – which eventually allows us to look into correspondence between biological and artificial systems –, we need to let human subjects recognize the same test images as deep neural networks. Yet, it is difficult to evaluate human object recognition performance if the number of object classes is too large. For example, it is complicated for a person to remember all ImageNet's 1000 classes to correctly recognize a given object. As indicated in [10], one source for human recognition error is the persons were not aware of the existence of the correct class.

Therefore, a dataset with a small number of classes may be more appropriate to compare classification accuracy between humans and deep neural networks. In [11], the authors test eight object classes using artificially distorted images. Similar works with five and ten categories are carried out in [12, 13]. Rajalingham et al. [14] utilize 24 categories; but only binary classification, i.e. making a choice between two available classes, is performed. While these studies allow us to evaluate visual recognition performance in a controlled context, e.g. noise or image variation, they do not consider widely used natural image datasets in the same way as tested with deep neural networks.

In the present work, we want to consistently evaluate human classification accuracy, and hence make a detailed comparison of recognition capability between humans and deep neural networks. As humans cannot remember a too large number of categories, we use CIFAR10, a common dataset of ten classes. There are several sources of difficulty in CIFAR10; its images represent a large variability in size, position, pose, and illumination. Human subjects will classify all the same test images as artificial recognition systems. This will set a classification accuracy reference for CIFAR10 and helps to interpret performance of recent successful deep neural networks, as well as understand their behavior.

The contributions of the present paper are as follows.
- According to our best knowledge, our work is the first to carry out an extensive experiment on a commonly used natural image dataset to consistently quantify human visual recognition ability and fairly compare it with deep neural networks' performance.
- We analyze in detail visual recognition performance of humans and deep neural networks according to multiple levels of difficulty. Different from previous works, these difficulty levels are based on human classification accuracy. This analysis helps to show strength and weakness of the state-of-the-art CNNs. Particularly, deep neural networks do not interpret recognition difficulty in the same way as humans; some images are hard for humans but less hard for CNNs, and vice versa.
- Our work reveals that while recent deep neural networks are highly effective in visual recognition, there is still room for their improvement. We indicate a subset of CIFAR10's test images, which are extremely easy for humans but requires further progress of CNNs to match human performance.
- While comparing deep neural networks with humans, we also present a general picture of the evolution of CNN-based visual recognition methods and, particularly, the core idea of residual learning and its variants that have brought about impressive success.

The rest of the present paper is as follows. Section 2 reviews previous studies on comparison of visual recognition performance between deep neural networks and humans. Section 3 summarizes and analyzes main CNNs used for object classification. We describe our experiment to evaluate human visual recognition performance in section 4. In section 5, we thoroughly compare visual recognition capability of deep neural networks with human performance. Discussion is mentioned in section 6. Finally, section 7 provides some conclusions from findings of our work.

## 2. Related work

Comparing neural networks' recognition performance with human capability is of great interest since it allows for evaluating artificial vision systems and can provide insights to improve them. In this section, we will focus on comparisons using deep neural networks, which have shown significant improvement in visual object recognition. For traditional machine learning algorithms, readers can see a thorough evaluation in [15].

In [10] Russakovsky et al. evaluate human object recognition performance on the ImageNet dataset and hence establish a reference for other works to improve deep CNNs' capability. ImageNet contains 1000 object categories. In the experiment, there are two human subjects; they are allowed to select five among 1000 categories for each presented image. Because of a large number of classes, the human subjects were trained with some validation images before the testing session. The participants were not under any time constraint. Yet it is observed that some images are quickly recognized, while harder images take multiple minutes for the human subjects to classify. The first person classified 1500 images, randomly selected from 100000 test images of the dataset. The top-5 error of this human subject is 5.1%. The other person recognized 258 images and obtained 12% for the top-5 error. Hence, the human error rate of 5.1% is then considered as human visual recognition performance and is widely used to compare with recognition models.

Other studies seem to focus on computer vs. human comparison for artificially modified images. Kheradpisheh et al. [12] show that deep CNNs can resemble human performance in invariant object recognition. This experiment considers five object categories; each of which has seven levels of variation. Variations are defined according to position, size, rotation in depth, and rotation in plane. The authors also take into account two kinds of background: uniform gray background and natural background. At the same time, ten models – eight of which are deep CNNs – are used for comparison. The results reveal that when variations are weak, shallow models can outperform deep networks and humans. In case of larger variations, deeper networks, i.e. more layers, are needed to match or even surpass human performance. However, it is worth noting that the number of images used to evaluate human performance is not the same as that for models. Specifically, human subjects view fewer images per object category than models.

Dodge and Karam [13] test a subset of ten classes from ImageNet; these ten classes concern only different categories of dogs. To generate distorted images, this work considers two types of distortion: additive Gaussian noise and Gaussian blur. In total, there are 1200 test images, including distorted and clean images. Participants are allowed to view images freely. Human subjects and neural networks predict images in the same procedure. The experiment shows that while CNNs can match or outperform humans on good quality images, CNNs' performance is much lower than human accuracy on distorted images. Moreover, there is no correlation in error between humans and CNNs.

Similar as [13], in [11] the authors evaluate human vs. computer recognition performance on distorted images, which are generated by adding noise and Gaussian blur. Yet the display time is limited to 100 ms to evaluate the early vision mechanism. Besides, the dataset is also different from [13]: there are eight object classes from Caltech101; it is considered easy dataset. Human subjects and models carry out the experiment in the same way. It is important to note that in [11, 13], human subjects do not view all test images: it is supposed that a subject who correctly recognizes an image with high level of distortion will also correctly classify a less distorted version. The results reveal that once again the human visual system outperforms deep neural networks. More results of the same tendency are observed with 16 object categories in [16].

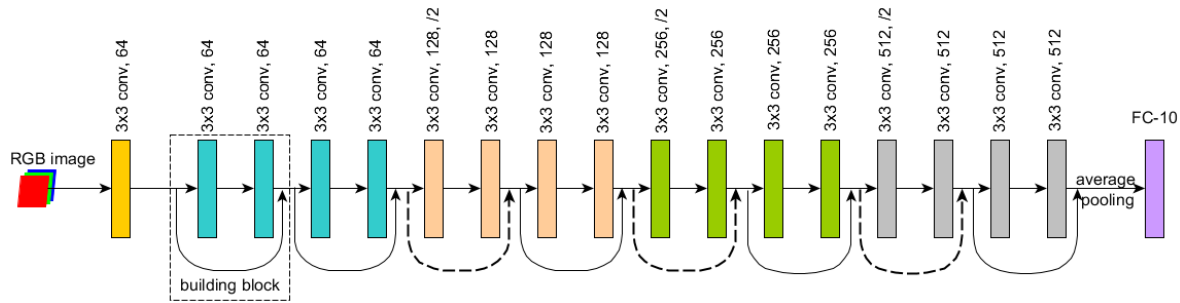

Figure 1: Example of ResNet's architecture [21]. The dashed shortcuts represent subsampling. The network has 18 weighted layers.

Another direction of comparison between deep neural networks and humans is to look at their recognition behavior and representation [9, 14, 17, 18]. In [14] the authors evaluate recognition performance – of humans and CNNs and monkeys – using 24 object classes and 2400 images. Yet, the task is only binary discrimination, i.e. given an image a subject or model has to choose an object between two possibilities. The results show that deep networks can predict human recognition behavior at object level or class level, but fail at individual image level.

In [9], the authors are interested in stage-wise computation. Visual brain representations are compared with those in deep neural networks. The work reveals hierarchical correspondence between artificial vision models and the human object recognition system.

While our work also aims at evaluating visual recognition capability of deep neural networks and humans, it focuses on natural images. In our experiment, human subjects classify the same test images as models. We want to look into classification performance of the state-of-the-art deep neural networks on CIFAR10, a dataset that has shown their strengths.

## 3. Deep neural networks for visual object recognition

Recently, many deep neural networks with various architectures have been proposed for computer vision, especially for visual recognition [5, 8, 19, 20, 21, 22]; although the first and basic convolutional neural network (LeNet) was presented several years before to recognize hand-written characters [23]. In this section, we will focus only on recent CNNs that dramatically improve object recognition accuracy. Generally, a CNN consists of several layers stacked upon each other; they are convolution (conv), Rectified Linear Unit (ReLU), Batch Normalization (BN) [24], pooling, dropout [25], and Fully-Connected layer (FC). Some basic layers can grouped together to create a block. FC layers are often used towards the end of a network.

A loss function between target (ground truth) and network's output is minimized to find the parameters or weights of the neural network. Gradient Descent with back propagation is almost exclusively used for optimization. For efficient convergence, some initialization and weight updating techniques have been proposed [7, 26].

### 3.1. VGG

VGG is one of the most well-known CNNs by introducing a new architecture and boosting recognition performance on the ImageNet dataset [27]. The particularity of VGG is all convolutional kernels are of size 3×3. All conv layers are followed by ReLU, while max-pooling is applied after some of them. VGG utilizes three FC layers at the end, meanwhile dropout is absent from the network.

When feature size is halved, width (i.e. number of feature maps or number of filters) is doubled to keep an equivalent complexity per layer. Another interpretation of such structure is as follows. After pooling to decrease feature size, the receptive field size increases and hence there are more variations in this image region, which in turn require more filters to represent. VGG is the first to show that a deep network (19 layers or more) is possible.

### 3.2. Inception/GoogLeNet

GoogLeNet (22 layers with parameter) is the implementation of the Inception architecture for ImageNet; it contains several Inception modules [28]. Each Inception module has four branches: three conv branches with kernel size of 1×1, 3×3, and 5×5, and one branch for max-pooling. These multi-branches represent multi-scale processing. Conv layers are always used with ReLU; dropout is near the network's output.

Besides, 1×1 conv layer is also utilized to reduce dimension (before 3×3 or 5×5 conv layers) or to match dimension (before addition). In fact, 1×1 conv layer is first proposed in the Network-in-Network architecture [29]. When combined with a traditional conv layer, it can enhance representational power while keeping the same receptive field size.

As in the VGG architecture, towards the network's output, feature size decreases and width increases.

### 3.3. ResNet

Residual network introduces a novel architecture that helps to deal with the degradation problem – higher training error when using more layers – and hence allows for training of a very deep network [8]. Since its appearance, ResNet has had a great impact on other deep networks' architecture and has dramatically increased visual object recognition performance.

The main idea of ResNet is residual learning: instead of approximating a mapping function $\{H(x)\}$ for some structure (containing one or several layers), it learns $\{H(x) - x\}$ and adds the identity $\{x\}$ at the output (figure 1).

ResNet consists of several stacked layers. Each layer contains three components in the following order: 3×3 conv, BN, and ReLU. Residual learning is applied for every two layers and creates a building block (or "non-bottleneck" building block). When shortcut (identity) combination exists, ReLU is applied after addition. No dropout is used in the network. In [8] He et al. also propose the "bottleneck" building block, in which instead of two 3×3 conv layers, they use three conv layers as follows: 1×1 conv – 3×3 conv – 1×1 conv.

Similar to previous networks, when feature size is halved, the number of filters (or width) is doubled. Particularly, across two different feature sizes, the shortcut (identity mapping) is transformed in either ways to match dimension: adding zeros (option A) or 1×1 conv (option B). Subsampling is carried out directly by convolution with stride 2.

Besides, there is one conv layer at the beginning of the network and one FC layer at the end. Before the FC layer, global average pooling is utilized. An interesting characteristic of ResNet is that it is highly modularized; and hence it can be easy to increase ResNet's depth. For example, in the case of CIFAR10, if there are four feature sizes {32, 16, 8, 4}, each size corresponds to 2n layers, where n represents the number of shortcuts (or "non-bottleneck" building blocks) per size [21]. Totally, the number of weighted layers is 8n+2 ("2" represents the first conv layer and the FC layer, ReLU and average pooling have no parameters, those of BN are negligible).

### 3.4. WideResNet

In [19], Zagoruyko and Komodakis thoroughly investigate residual network's architecture proposed in [8], particularly depth (number of layers) versus width (number of feature maps or filters). The results of this work show that wide residual networks (WRN) present advantages over thin and deep counterparts. WRN can be trained faster and provide the state-of-the-art visual recognition performance. Besides, it also shows that it is residual learning, and not extreme depth, that really brings about CNNs' power.

Like residual network in [8], WRN consists of block groups, each of which has some building blocks. Parallel to the shortcut, the residual branch of a building block contains two 3×3 conv layers. The conv layer's order is BN-ReLU-Conv, which is different from the Conv-BN-ReLU order in the original residual network. Another important difference is the use of dropout in WRN. Due to the increased number of parameters in a layer, dropout is inserted between two conv layers in the residual branch. Dropout is proved to increase WRN's performance.

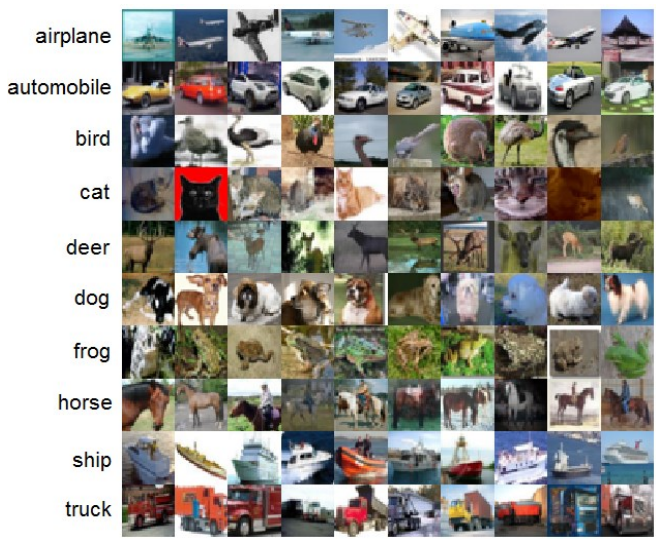

Figure 2: Example of ten categories of CIFAR10.

### 3.5. ResNeXt

Xie et al. [20] continue to test a new dimension in residual network, called cardinality, which is the number of residual branches. Different from ResNet or WRN, which has only one residual branch, ResNeXt utilizes many identical branches (for example, 32 branches or cardinality = 32). This idea is similar to Inception's architecture, but Inception has much fewer branches and they are not identical. ResNeXt exploits the bottleneck building block's architecture, in which a layer's order is conv-BN-ReLU, and does not utilize dropout as in [8].

"Shake-shake" network [22] also uses multiple residual branches, each of which is multiplied by a random number. This represents a regularization effect as dropout and can be applied in both forward and backward directions.

## 4. Experiment

### 4.1. Dataset

CIFAR10 is a dataset of natural color images, which is widely used to evaluate recognition capability of deep neural networks [8, 19, 20, 21, 30]. This dataset contains 60000 small images of 32×32 pixels from ten categories (or classes): airplane, automobile, bird, cat, deer, dog, frog, horse, ship, and truck. It is divided into two sets: training set of 50000 images and test set of 10000 images (1000 images per category). The images are taken under varying conditions in position, size, pose, and illumination. Many objects are partially occluded. Figure 2 illustrates ten images for each category of CIFAR10.

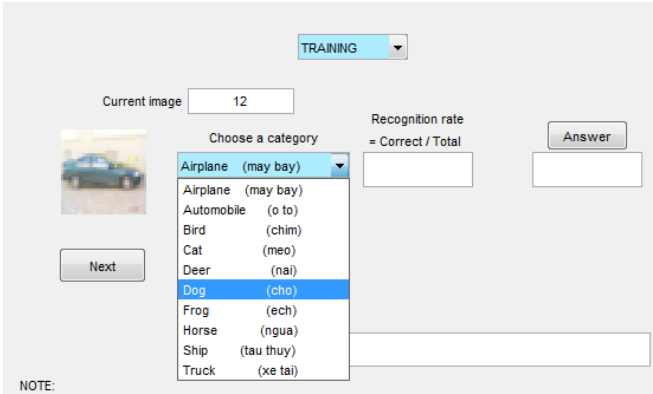

Figure 3: Interface used for image classification in the experiment. In the testing step, participants never know the ground truth label of a test image (even after it has been classified).

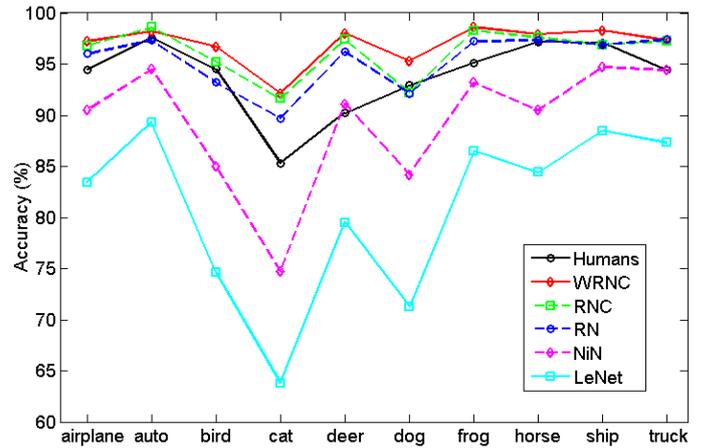

Figure 4: Recognition performance per category.

| Model | Humans | WRNC | RNC | RN | NiN | LeNet |
|---|---|---|---|---|---|---|
| Acc (%) | 93.91 ±1.52 | 96.96 | 96.18 | 95.33 | 89.28 | 80.86 |

Table 1: Recognition performance on all test images of CIFAR10

### 4.2. Procedure

We set up an experiment of CIFAR10 recognition in which human subjects carry out the classification task in the same way as deep neural networks. This task is realized through a Matlab-based interface (figure 3). When an image is displayed, a human subject selects one category that he/she thinks best matches the image.

The experiment consists of two steps. In the first step, participants are asked to train themselves to be familiar with the ten categories, which are displayed sequentially. They can use as many from the 50000 training images as possible for the training purpose. Yet we observed that participants did not view all 50000 images because normally they had already been familiar with these image categories. In the second step, each human subject classifies 1000 out of the 10000 test images.

It is very complicated for a person to classify all 10000 images; hence, we need multiple persons to complete these images and obtain a consistent evaluation of human recognition performance on CIFAR10. The 10000 test images are divided into ten groups of 1000 images: group 1 consists of test images numbered from 1 to 1000, according to the dataset provided by [30]; group 2 contains images 1001 to 2000, and so on. A human subject recognized images from only one group, i.e. 1000 images. It is important to note that these ten groups do not correspond to ten image categories: each group contains images from ten categories.

Sixty students participated in this experiment[1], six students per group. In other words, each image from the 10000 test images was recognized by six human subjects.

There is no time constraint for training and testing as in [10]. To avoid participants' fatigue, during the testing phase, they were allowed to make a pause after every 50 images. Hence, the time for a participant to finish the task may span a few days. It is observed that generally participants recognize an image quickly, except for harder images they need more time to classify.

### 5. Results

We now compare human recognition performance obtained from the experiment in section 4 with performance of a representative subset of deep neural networks. We choose a variety of CNNs from the basic network LeNet [23] to the state-of-the-art residual networks [8, 19]. In [21], the authors show that cutout regularization – i.e. randomly remove a patch in a training image – can help to improve classification accuracy. Thus, we also tested this technique with existing residual networks and observed significant improvements. Since VGG and GoogLeNet's classification performance is now surpassed by that of recent residual networks, we do not include them in our comparison. Finally, the following five networks are selected for comparison: LeNet, Network-in-Network (NiN), Residual network (RN), Residual network combined with cutout regularization (RNC), and Wide residual network combined with cutout regularization (WRNC). The architecture of RN is illustrated in figure 1. All these networks, implemented as in [21, 31], were trained until convergence.

It is important to note that both humans and CNNs classify all 10000 test images of the CIFAR10 dataset.

### 5.1. All test images

---

[1] Some students did not correctly follow the requirements of the experiment and, therefore, their data were rejected. The experiment's data and the easy subset of the CIFAR10 test dataset (section 5.2) can be downloaded at https://sites.google.com/site/hophuoctien/projects/virec/cifar10-classification

| Level of difficulty | 1 | 2 | 3 | 4 | 5 | 6 | 7 |
|---|---|---|---|---|---|---|---|
| Number of images | 7929 | 1247 | 398 | 212 | 121 | 64 | 29 |
| Average No. images/category | 793±87 | 125±29 | 40±26 | 21±18 | 12±13 | 6±6 | 3±4 |

Table 2: Levels of difficulty based on human classification accuracy.

The average recognition accuracy of all human subjects is 93.91±1.52%. If we consider only the ten highest group rates (one per group), the average accuracy is 95.78±0.78. From table 1, we can see that recent CNNs, particularly WRNC, have much improved recognition capability and have already surpassed human performance. These results also confirm the role of residual learning, which is also the basic part of modern deep networks, in object recognition. It is not surprising that LeNet provides the lowest accuracy since it is very simple and is originally designed for hand-written character recognition.

Figure 4 represents performance of humans and CNNs for different categories. This result reveals that some category is harder to recognize than others. Besides, the accuracy curves of all the models and humans show similar tendency. For example, when humans obtain the lowest accuracy with cat images, this situation also appears in all the five CNNs.

At category level, human performance is also better than NiN and LeNet but is not as good as recent deep neural networks. We observe that while there is variation in performance between categories, successful CNNs reduce such difference.

### 5.2. Levels of difficulty

In order to know more in detail about deep neural networks' capability, we look into their classification accuracy according to difficulty level of CIFAR10's test images. Different from previous studies [11, 13, 16], our work defines recognition difficulty based on human subjects. Thus, the test dataset is divided into seven levels of difficulty. The first level consists of images that all human subjects recognized correctly. The second level contains images that were wrongly classified by only one person (one among six persons classifying an image). The third level corresponds to images wrongly recognized by exactly two persons. Similarly, no subjects correctly recognized the images of the highest level of difficulty (level 7).

The number of images per difficulty level is given in table 2. The number of images decreases when level of difficulty increases. For instance, level 1 consists of 7929 images; meanwhile, level 7 has only 29 images. Table 2 also represents the average number of images per category. Except the highest difficulty level that has only six categories, all other levels contain images from ten categories.

Figure 5 shows recognition performance of humans and deep neural networks according to levels of difficulty. Human performance logically decreases from 100% at level 1 to 0% at level 7, and is considered as a baseline to compare with deep neural networks. It is interesting to note that almost all CNNs perform better than humans at high levels of difficulty. Specifically, a simple model like LeNet, which uses only five convolutional layers, recognizes visual objects better than humans when difficulty level is higher than 3. Between CNNs, we observe the same tendency as in section 5.1: WRNC provides the best performance and LeNet the worst. The performance of all these models decreases according to levels of difficulty, although the rate of decrease is less than that of humans.

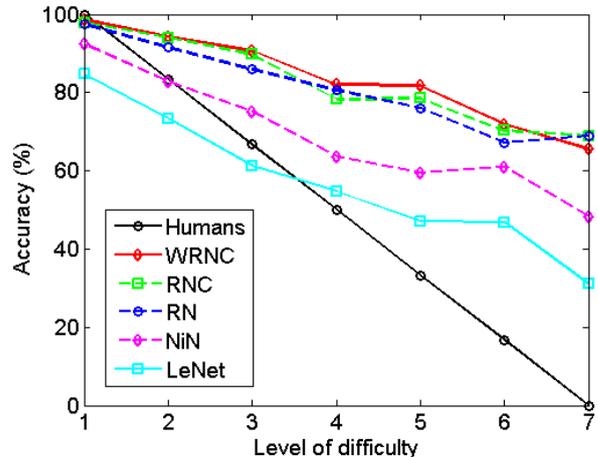

Figure 5: Recognition performance according to difficulty levels

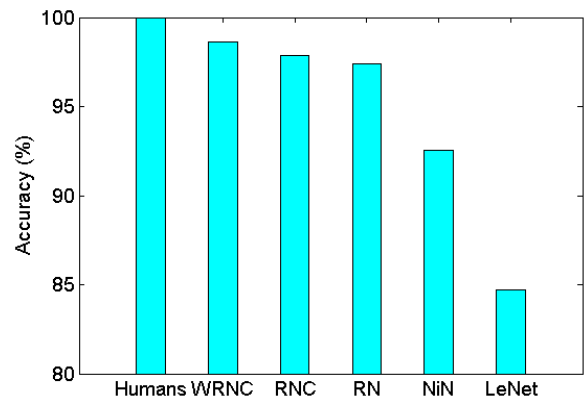

Figure 6: Recognition performance at the first level of difficulty.

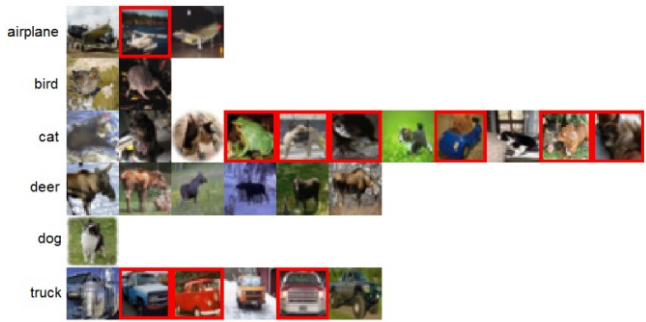

Figure 7: The images that no human subjects recognized correctly. The ones with red box are those that WRNC recognized wrongly.

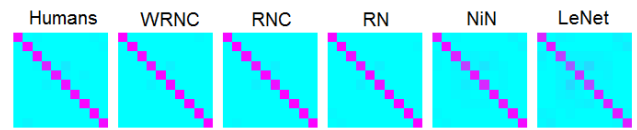

Figure 8: Confusion matrices of humans and CNNs with all test images. Color code is the same as in figure 9

However, no CNNs can match the perfect accuracy (100%) of humans at level 1 (figure 6). The best CNN (WRNC) obtains 98.64% accuracy, followed by RNC (97.89%) and RN (97.40%). In fact, level 1 contains a large number of test images (7929), which are clear and so easy for humans to recognize. We call level 1 the easy subset of CIFAR10. Our result suggests that deep neural networks are still to be improved to be as reliable as humans in object recognition. In that case, the easy subset of CIFAR10 may be useful for evaluating future deep networks.

Figure 7 illustrates the hardest images for humans. It seems that hard images are not really due to degradation of quality such as blur or noise, but related to confusion between categories, for example between cat and dog, deer and horse. CNNs, for example WRNC in figure 7, seem to be less prone to this kind of error. One explanation may be that humans pay more attention to overall information and try to understand the meaning of a scene. Hence, when there is confusion, humans are likely to make error. Meanwhile, since neural networks focus more on detail and may predict a category simply by searching correlation between a test image and those they have learnt, this increases chance of correct classification.

### 5.3. Confusion matrix

Confusion matrix allows us to see the distribution of recognition error. Each row of a 10×10 confusion matrix represents a ground truth category; each column indicates the probability that a category is chosen given the ground truth. The sum of each row is equal to one. Figure 8 shows the confusion matrices of humans and CNNs using all test images. The results of WRNC, RNC, and RN are quite similar and are closed to human behavior. As imagined, NiN and LeNet have slightly different confusion matrices since they provide lower recognition accuracy.

Similarly, figure 9 illustrates the confusion matrices of humans and CNNs for multiple levels of difficulty. Level 7 is not displayed because some categories do not contain any image at this level. At level 1, the human confusion matrix is diagonal. At higher levels, it increasingly loses this structure. As in section 5.2, the confusion matrices of CNNs are not as perfect as the human confusion matrix at the first level of difficulty, but they look better than the human matrix at higher

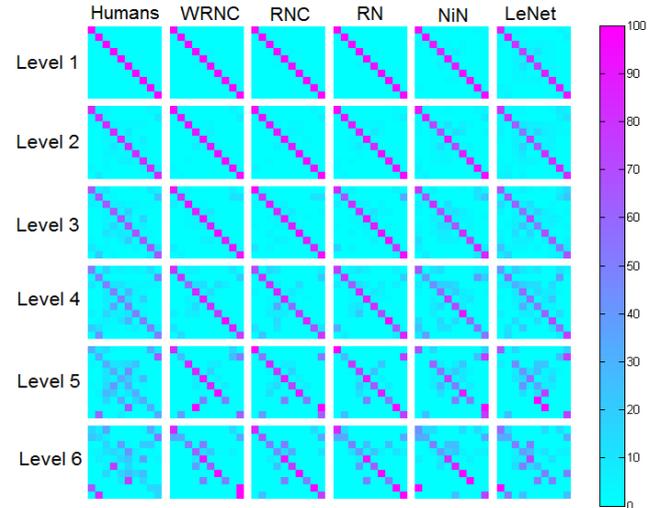

Figure 9: Confusion matrices according to levels of difficulty. Level 1 represents the easiest level for humans.

levels. Besides, the confusion matrices of WRNC, RNC and RN share more or less similarity.

## 6. Discussion

**Deep neural networks' recognition performance**. Recent deep neural networks have outperformed humans in visual object recognition on the CIFAR10 dataset, as reported on other datasets such as ImageNet [7]. Yet more improvements are required and many issues still need to be illuminated. While the state-of-the-art CNNs can perform better than humans on images that humans find hard, they do not match human performance on easy images. As we see in the experiment, about 80% of CIFAR10's test images are easy and humans can correctly recognize them without effort. In fact, it is observed that human subjects classified these images very quickly. That means neural networks are not as reliable as humans and hence there is still room for recognition models to improve their performance in visual object classification. In that case, the easy subset of CIFAR10 defined by level 1 in the present work may be helpful for evaluating future deep networks' recognition capability.

The fact that CNNs are able to classify objects better than humans for hard images and worse for easy ones, reveals that it may be still far from affirming that deep neural networks can understand the meaning of scene, or at least understand images in the same way as humans. It is probable that deep neural networks recognize an image by exploiting similarity or correlation between it and other images they have learned. Such similarity search may be purely computational and does not suppose a capability of scene understanding or generalization.

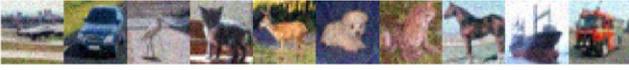

Figure 10: Example of noisy images with Gaussian noise ($\sigma = 0.05$, pixel values are between 0 and 1).

| Accuracy (%) | RNC | WRNC | Humans |
|---|---|---|---|
| $\sigma = 0.03$ | 77.35 % | 84.40 % | X |
| $\sigma = 0.05$ | 52.06 % | 60.90 % | 90.10% |

Table 3: Recognition performance on noisy images.

To test generalization capability of above deep neural networks, we evaluate them on noisy images. These images are simply 10000 test images of CIFAR10 corrupted by Gaussian noise with zero mean and variance $\sigma^2$ (figure 10). The best two networks in section 5, i.e. RNC and WRNC, which are trained with CIFAR10's training images, are used to classify the noisy images. Table 3 shows that recognition performance of the two CNNs decreases significantly. For a rough estimation of human performance on this new dataset, another human subject, who had not participated in the experiment, was asked to recognize 1000 noisy images (among 10000 and only with $\sigma = 0.05$). This person obtained much higher accuracy than CNNs[2]. Our result is in line with previous studies dealing with degraded images [11, 13]. It also corroborates the conclusion in [32] that it is very easy to fool a deep neural network.

It is likely that deep neural networks will perform better on noisy images if they are trained with this kind of images. Nevertheless, we can also say that even the state-of-the-art deep neural networks have not possessed genuine capability of generalization, and thus much more effort should be done for them to attain this human-like capability. Neural networks need to be able to simulate mechanism guiding scene understanding to robustly recognize objects though they may appear in noisy environment or even in unusual conditions such as abnormal color [33].

It is known that an effective visual recognition system should take into account invariance transformations concerning position, pose, size, and illumination. CIFAR10 allows for considering these transformations in an uncontrolled manner. While some studies disclose deep neural networks' capability of learning invariance transformations [12], the performance of the state-of-the-art CNNs on easy images (level 1) in our experiment suggests that this characteristic may still be improved to match the human level. Comparing CNN layers and responses at the neuronal level in the human visual system as in [9] may help explaining the mechanism of human visual recognition and hence improving neural networks. Furthermore, though all the state-of-the-art CNNs are feedforward, existing studies show recurrent connections may play a certain role in object recognition [34, 35]. Future CNNs may imitate such architecture for better performance. In parallel, it is worth paying more attention to interpretable deep neural networks as this direction of research has shown encouraging results [36, 37].

**Level of difficulty.** Other works in the literature degrade images' quality and then compare recognition performance on these images between humans and deep neural networks [11, 13, 16]. These studies consider multiple levels of difficulty, which are based on stimulus quality using factors such as noise or blur. In contrast, in our work, the levels of difficulty are defined based on human recognition performance, and are employed to reveal strength and weakness of the state-of-the-art deep networks. Thus, while in previous studies, deep neural networks can match human performance for clean or good quality images and fail to do so for low quality images; our work shows that deep networks outperform humans for hard images and do not match human capability for easy ones. Our work is complementary to previous studies in helping to further understand the behavior of humans and artificial vision models in object classification. It confirms the divergence in recognition behavior between the two systems, which has been revealed in other researches [14, 38]. Hence, since human visual recognition strengths are not the same as those of deep neural networks, it will be interesting to exploit these different strengths in order to improve artificial recognition systems.

**CIFAR10.** Since CIFAR10 consists of tiny images, the present work provides a human performance reference for recognition of low resolution natural scene. This kind of image may appear in many situations. In fact, the output of some image sensors, e.g. THz imagers, is very low resolution [39]. Some applications also require recognition of tiny images. For example, in order to count the number of objects or faces in a large picture, we may recognize small objects before counting them [40].

Low resolution may also influence human recognition performance. Using CIFAR10's tiny images can increase the chance that human subjects confuse a category with one another, for example deer vs. horse in the experiment. The human performance in the present work is close to that of [10], which is based on larger images and a much higher number of classes. It is expected that human performance would have been better for recognition of ten classes if we had used larger images. Besides, while the ten categories of CIFAR10 are well known to human subjects, there are also hard images. Although participants were asked to train themselves with as many images as possible, they were likely to stop training early since they often saw easy images. If participants use more training time for hard images, they can increase their classification accuracy. Especially, because of the low resolution of CIFAR10's images, correct recognition of some hard images may be related to similarity exploitation between training and test images rather than understanding scene meaning. Yet this might favor deep neural networks, which are trained from all the training dataset. To confirm this hypothesis, we would need to carry out further investigation on similarity between CIFAR10's training and test images, particularly for hard ones.

---

[2] At the time of writing the present paper, "Shake-Shake" network [22] gives somewhat better accuracy than WRNC, but it is highly likely that it does not change our conclusion.

## 7. Conclusion

The present paper compares visual recognition performance of deep neural networks with human ability on CIFAR10, a natural image dataset widely used for evaluation of object classification models. The comparison is carried out on a level playing field: both humans and deep networks classify the same test images. CIFAR10-based evaluations reveal both strength and weakness of the state-of-the-art deep neural networks.

By considering the whole dataset, recent deep neural networks surpass humans in visual recognition. Yet detailed investigation discloses difference in behavior between the two systems of recognition. To highlight this divergence, our work evaluates classification performance of humans and CNNs according to multiple difficulty levels. These levels of difficulty are based on human classification accuracy. The results show that easy images for humans may still not be easy for deep neural networks. Such kind of images can be used to improve future recognition models.

It is likely that the human visual system can provide insights to build more effective neural networks. One direction may be to simulate in deep networks human abilities of scene understanding and generalization, which would eventually help them to recognize images more robustly.

## References

[1] J. J. DiCarlo, D. Zoccolan, and N. C. Rust, "How does the brain solve visual object recognition?," *Neuron*, vol. 73, no. 3, pp. 415-434, 2012.
[2] Y. LeCun, Y. Bengio, and G. Hinton, "Deep learning," *Nature*, vol. 521, pp. 436–444, 2015.
[3] J. Schmidhuber, "Deep learning in neural networks: An overview," *Neural Networks*, vol. 61, pp. 85-117, 2015.
[4] Y. Bengio, A. Courville, and P. Vincent, "Representation Learning: A Review and New Perspectives," *PAMI*, vol. 35, no. 8, pp. 1798-1828, 2013.
[5] A. Krizhevsky, I. Sutskever, and G. E. Hinton, "ImageNet classification with deep convolutional neural networks," in *NIPS*, 2012, pp. 1097-1105.
[6] Y. Bengio, "Learning Deep Architectures for AI," *Foundations and Trends in Machine Learning*, vol. 2, no. 1, pp. 1-127, 2009.
[7] K. He, X. Zhang, S. Ren, and J. Sun, "Delving Deep into Rectifiers: Surpassing Human-Level Performance on ImageNet Classification," in *ICCV*, 2015.
[8] K. He, X. Zhang, S. Ren, and J. Sun, "Deep Residual Learning for Image Recognition," in *CVPR*, 2016.
[9] R. M. Cichy, A. Khosla, D. Pantazis, A. Torralba, and A. Oliva, "Comparison of deep neural networks to spatio-temporal cortical dynamics of human visual object recognition reveals hierarchical correspondence," *Scientific Reports*, vol. 6, no. 27755, 2016.
[10] O. Russakovsky et al., "ImageNet Large Scale Visual Recognition Challenge," *International Journal of Computer Vision*, vol. 115, no. 3, pp. 211–252, 2015.
[11] S. Dodge and L. Karam, "Can the early human visual system compete with Deep Neural Networks?," in *ICCV*, 2017, pp. 2798-2804.
[12] S. R. Kheradpisheh, M. Ghodrati, M. Ganjtabesh, and T. Masquelier, "Deep networks can resemble human feedforward vision in invariant object recognition," *Scientific reports*, vol. 6, no. 32672, 2016.
[13] S. Dodge and L. Karam, "A Study and Comparison of Human and Deep Learning Recognition Performance under Visual Distortions," in *International Conference on Computer Communications and Networks*, 2017.
[14] R. Rajalingham et al., "Large-scale, high-resolution comparison of the core visual object recognition behavior of humans, monkeys, and state-of-the-art deep artificial neural networks," *Neurosci*, vol. 38, no. 33, pp. 7255-7269, 2018.
[15] A. Borji and L. Itti, "Human vs. computer in scene and object recognition," in *CVPR*, 2014.
[16] R. Geirhos et al., "Comparing deep neural networks against humans: object recognition when the signal gets weaker," *CoRR*, 2017.
[17] N. Pinto et al., "Human versus machine: comparing visual object recognition systems on a level playing field," in *Computational and Systems Neuroscience*, 2010.
[18] C. F. Cadieu et al., "Deep neural networks rival the representation of primate IT cortex for core visual object recognition," *PLoS computational biology*, vol. 10, no. 12, 2014.
[19] S. Zagoruyko and N. Komodakis, "Wide residual networks," in *BMVC*, 2016.
[20] S. Xie, R. Girshick, P. Dollár, Z. Tu, and K. He, "Aggregated Residual Transformations for Deep Neural Networks," in *CVPR*, 2017.
[21] T. DeVries and G. W. Taylor, "Improved Regularization of Convolutional Neural Networks with Cutout," *arXiv*, 2017.
[22] X. Gastaldi, "Shake-Shake regularization," *arXiv*, 2017.
[23] Y. LeCun, L. Bottou, Y. Bengio, and P. Haffner, "Gradient-based learning applied to document recognition," *Proceedings of the IEEE*, vol. 86, no. 11, pp. 2278–2324, 1998.
[24] S. Ioffe and C. Szegedy, "Batch Normalization: Accelerating Deep Network Training by Reducing Internal Covariate Shift," in *ICML*, 2015.
[25] N. Srivastava, G. Hinton, A. Krizhevsky, I. Sutskever, and R. Salakhutdinov, "Dropout: a simple way to prevent neural networks from overfitting," *The Journal of Machine Learning Research*, vol. 15, no. 1, pp. 1929-1958, 2014.
[26] D. Kingma and J. Ba, "Adam: A Method for Stochastic Optimization," in *International Conference on Learning Representations (ICLR)*, 2015.
[27] K. Simonyan and A. Zisserman, "Very Deep Convolutional Networks for Large-Scale Image Recognition," *arXiv*, 2014.
[28] C. Szegedy et al., "Going deeper with convolutions," in *CVPR*, 2015.
[29] M. Lin, Q. Chen, and S. Yan, "Network in network," *CoRR*, 2013.
[30] A. Krizhevsky, "Learning Multiple Layers of Features from Tiny Images," *Technical report*, 2009.
[31] A. Vedaldi and K. Lenc, "MatConvNet: Convolutional Neural Networks for MATLAB," in *ACM international conference on Multimedia*, 2015, pp. 689-692.
[32] A. Nguyen, J. Yosinski, and J. Clune, "Deep Neural Networks are Easily Fooled: High Confidence Predictions for Unrecognizable Images," in *Computer Vision and Pattern Recognition*, 2015.
[33] T. Ho-Phuoc, N. Guyader, F. Landragin, and A. Guérin-Dugué, "When viewing natural scenes, do abnormal colors impact on

spatial or temporal parameters of eye movements?," *Journal of Vision*, vol. 12, no. 2, pp. 1-13, 2012.
[34] R. C. O'Reilly, D. Wyatte, S. Herd, B. Mingus, and D. J. Jilk, "Recurrent Processing during Object Recognition," *Front Psychol*, vol. 4, no. 124, 2013.
[35] J. Donahue et al., "Long-term recurrent convolutional networks for visual recognition and description," *PAMI*, vol. 39, no. 4, pp. 677 – 691, 2017.
[36] S. Mallat, "Understanding deep convolutional networks," *Philos Trans A Math Phys Eng Sci.*, vol. 374, no. 2065, 2016.
[37] D. Bau, B. Zhou, A. Khosla, A. Oliva, and A. Torralba, "Network Dissection: Quantifying Interpretability of Deep Visual Representations," in *CVPR*, 2017.
[38] S. Ullman, L. Assif, E. Fetaya, and D. Harari, "Atoms of recognition in human and Computer vision," *PNAS*, vol. 113, no. 10, pp. 2744-2749, 2016.
[39] A. Boukhayma, A. Dupret, J. P. Rostaing, and C. Enz, "A Low-Noise CMOS THz Imager Based on Source Modulation and an In-Pixel High-Q Passive Switched-Capacitor N-Path Filter," *Sensors*, vol. 16, no. 3, 2016.
[40] Y. Bai, Y. Zhang, M. Ding, and B. Ghanem, "Finding Tiny Faces in the Wild with Generative Adversarial Network," in *Computer Vision and Pattern Recognition*, 2018.